\documentclass[12pt,a4paper]{article}

\usepackage[a4paper,width=150mm,top=25mm,bottom=25mm]{geometry}
\usepackage[utf8]{inputenc}

\usepackage{graphicx}
\usepackage{placeins}
\usepackage{textcomp}
\usepackage{xcolor}
\PassOptionsToPackage{hyphens}{url}
\usepackage[
  colorlinks=true,
  urlcolor=blue,
  linkcolor=blue,
  citecolor=blue,
]{hyperref}

\usepackage{booktabs, dcolumn}
\def\sym#1{\ifmmode^{#1}\else\(^{#1}\)\fi}
\usepackage[flushleft]{threeparttable}

\usepackage{subfig}
\usepackage{ragged2e}

\usepackage{authblk}
\usepackage{orcidlink}

\usepackage{amsmath,amssymb,amsfonts}
\usepackage{algorithmic}
\usepackage{bbm}

\usepackage{comment}
\usepackage{float}
\usepackage[
    backend=biber,
    style=authoryear,
    citestyle=apa,
    isbn=false,
    url=false,
    date=year,
    clearlang=true
]{biblatex}
\AtEveryBibitem{\clearlist{language}}
\AtEveryBibitem{\clearlist{version}}

\newcommand{\analysisNFamilies}{2,855,109}
\newcommand{\analysisNLinks}{15,943,404}
\newcommand{\analysisNMatch}{2,150,780}
\newcommand{\analysisNPapapers}{4,812,869}

\title{Tracing the Flow of Knowledge From Science to Technology Using Deep Learning\thanks{Our work has been enabled by grant 2023/8401 from the European Patent Office under their Academic Research Programme (EPO-ARP) 2021. The project has benefited from discussions with Fabian Gaessler, Karin Hoisl, Matt Marx and Felix Poege. We furthermore thank participants at the NBER I3 Technical Working Group Meeting in Fall 2024 in Cambridge, MA as well as EPIP 2025 in Antwerp for helpful feedback and suggestions. Finally, we also thank seminar participants at Polytechnic University of Milan, Manchester University (Alliance Manchester Business School) and University of Sussex for helpful comments.}}
\date{}
\author[1]{Michael E. Rose \orcidlink{0000-0002-4128-4236}}
\author[1]{Mainak Ghosh \orcidlink{0000-0003-1176-4236}}
\author[1]{Sebastian Erhardt \orcidlink{0000-0002-2933-6451}}
\author[1]{Cheng Li \orcidlink{0000-0002-9016-5874}}
\author[1]{Erik Buunk \orcidlink{0000-0003-1604-0421}}
\author[1]{Dietmar Harhoff \orcidlink{0000-0001-7708-292X}}
\affil[1]{Max Planck Institute for Innovation and Competition, Munich, Germany}

\begin{document}

\maketitle
\thispagestyle{empty}

\begin{abstract}
We develop a language similarity model suitable for working with patents and scientific publications at the same time. In a horse race-style evaluation, we subject eight language (similarity) models to predict credible Patent-Paper Citations. We find that our Pat-SPECTER model performs best, which is the SPECTER2 model fine-tuned on patents. In two real-world scenarios (separating patent-paper-pairs and predicting patent-paper-pairs) we demonstrate the capabilities of the Pat-SPECTER.\\
We finally test the hypothesis that US patents cite papers that are semantically less similar than in other large jurisdictions, which we posit is because of the duty of candor. The model is open for the academic community and practitioners alike.
\end{abstract}


% TEXT
\newpage
\setcounter{page}{1}
\section{Introduction}\label{ch:introduction}

The translation of science to industrial applications is pivotal for modern innovation processes. Firms rely on knowledge generated in the public sector \parencite{aghion_academic_2008}, yet inventors face an enormous search and attention problem when engaging with academic literature \parencite{bikard_made_2018}.

At the same time, scholars and evaluators are interested in tracing the flow of knowledge from science to technology to understand innovation processes. The dominant approach for tracing knowledge flows is the analysis of explicit citations in scientific publications and patents. Such citations acknowledge intellectual priority and delineate the foundations upon which new inventions are built. This method, however, has significant limitations, such as the sparsity of citation graphs and the strategic or opportunistic selection of citations.

To overcome both problems, we present a transformer-based similarity model that can be used for patent-publication comparisons. We identify four suitable transformer models and test them rigorously in a horse race. These models include SPECTER, developed by \textcite{cohan_specter_2020} for work with publications, and PaECTER, developed by \textcite{ghosh_paecter_2024} for work with patents, as well as derivations from these two models. We benchmark the models using standard test metrics on small datasets derived from Patent Paper citations by \textcite{marx_reliance_2022,marx_patent-paper_2023}.

The winning model, Pat-SPECTER, is the SPECTER model fine-tuned on patents. The open-source model generates embeddings that allow semantic comparisons of English publications and patents at the same time.

To validate the Pat-SPECTER, we apply the model to two real-world scenarios, namely the separation of patent-paper pairs from random patent-to-paper pairings, and the  prediction of patent paper pairs among millions of possible documents. We utilize an ElasticSearch database in the Logic Mill search system by \textcite{erhardt_logic_2024} to identify (approximate) nearest neighbors among all of PATSTAT and OpenAlex.

To show the model's strength, we finally investigate a question at the core of innovation research: when does transfer from science to technology take place? More specifically, we investigate when citations to publications on patents are semantically similar to the patent. We find that in jurisdictions that require applicants to perform a 'duty of candor', cited papers tend to be less related to the focal patent. The duty of candor includes an affirmative obligation to disclose all known material prior art, and violations can render a patent unenforceable. The duty of candor is a functional feature of the USPTO but also at the Israeli Patent Office (where it is called 'Affirmative Duty of Disclosure').

\textcite{michel_patent_2001} have argued that the USPTO's duty of candor induces applicants to cite more than is necessary -- a hypothesis that  \textcite{cotropia_applicant_2013} have tested and confirmed quantitatively by asking whether patent examiners actually use applicant-provided citations. Their answer is negative. Recently, \textcite{tur_extreme_2024} argue that the duty of candor leads applicants to cite a large amount of references, especially after a company sees its patents challenged in court. Though neither study distinguishes by type of cited document (patent or paper or else), the insights suggest that mandatory disclosure of prior art tends to produce irrelevant citations. By testing our model, we thus add a qualitative dimension to the existing quantitative analysis.

The motivation for our research is a lack of tools for innovation economists. In the patent domain, citations to scientific literature are used to link patented inventions to their scientific underpinnings. As many patents do not have direct references to the non-patent literature, \textcite{ahmadpoor_dual_2017} propose the notion of “distance to the science frontier”. 
Patents citing scientific publications directly are taken to be at the science frontier. 

\textcite{poege_science_2019} show that highly cited scientific publications often serve as the foundation for particularly impactful patented inventions, highlighting the critical role of science in driving technological innovation. \textcite{poege_science_2019} linked around 950,000 patent families to over 2.2 million scientific articles, demonstrating that foundational scientific research significantly contributes to breakthrough inventions.
Related studies such as \textcite{marx_reliance_2020} also use large-scale citation graphs to derive conclusions for innovation policy.

However, the use of explicit citations has serious disadvantages, even if indirect links are taken into account. Citations are rare, and they may be selected strategically or opportunistically \parencite{jaffe_patent_2017}.

As an alternative to citations, scholars have investigated the possibility of using text as input data and comparing two documents to gauge their similarity. 
When vectors represent individual documents, vector operations (such as Cosine distance, Manhattan distance, Euclidean distance or others) yield similarity estimates.

In his dissertation, \textcite{natterer_entwicklung_2014} uses a term frequency–inverse document frequency (TF-IDF) model of textual similarity for technical texts. A more recent example is \textcite{kelly_measuring_2021}, who identify breakthrough patents as patents with low textual similarity to the existing patent text corpus. They then establish that sectors with many breakthrough patents experience higher growth. This relationship could not have been demonstrated convincingly using patent citations, as the authors point out.

Yet, turning massive amounts of text into data is challenging, and so far there exists no scalable solution that at the same time spans diverse text corpora. All applications intended to find similarities between patents or scientific publications suffer from one or several of the following shortcomings: they are specific to only one text corpus (i.e., only patents, such as patent maps); they do not account for semantic structure across different text corpora; they do not scale well.

A common yet simple approach to transform text into a vector is the bag-of-word approach. Typically researchers used a weighted incarnation of this approach, the so-called term frequency–inverse document frequency (TF-IDF).

TF-IDF vectorization, simple as it is, has two important limitations. First, it ignores the relative positioning of terms (to each other and within the document) and scales badly. For example, including new documents in the corpus may require the re-computation of the entire matrix. Thus, it becomes computationally expensive with the growing number of documents.  Secondly, the TF-IDF matrix is sparse and high-dimensional, which leads to higher memory consumption. The loss of information of the location of a term within a sentence, within a paragraph and within a document is presumably the most severe limitation.

Recent advances in Natural Language Processing, especially the Word2vec \parencite{mikolov_efficient_2013}, and Bidirectional Encoder Representations from Transformers (BERT) \parencite{devlin_bert_2019}, have made it possible for non-computer scientists to work with textual data while retaining full syntactical information.

The BERT model leverages the attention layer \parencite{vaswani_attention_2017}, which efficiently estimates which tokens of a sentence to what extent are valuable in understanding the sentence. In this family of models, words in a sentence do not have equal weight, as is the case for TF-IDF. BERT is pre-trained on newspaper articles and Wikipedia and focuses on the 40,000 most important tokens.\footnote{A token usually represents a word or a part of a word.}

To date, many thousands of specialized transformers exist. Two notable models are SciBERT \parencite{beltagy_scibert_2019} and BERT for patents \parencite{srebrovic_leveraging_2020}. SciBERT is a BERT model specific to scientific publications. It has learned the 40,000 most important tokens used in scientific publications. In the patent domain, Google’s BERT for patents was trained on 100 million granted USPTO patents.

The computational requirements to train a BERT from scratch are immense, however. \textcite{beltagy_scibert_2019} state on their SciBERT paper that 16 TPUv3 (Tensor Processing Units of the third generation) chips were used for 4 days to train BERT’s largest model - 8 GPUs (Graphics Processing Unit) are expected to take 40-70 days for the same task. Training a base model can become very expensive and time-consuming. Thus, in our approach, we fine-tune existing models, as this is more efficient and allows us to leverage the knowledge already incorporated in the more fundamental models. 

Since BERT’s token vocabulary is likely not representative of every domain, multiple domain-specific BERT derivatives have been trained. For example, SciBERT, whose 30k most important tokens overlap with the original general purpose BERT’s vocabulary at the rate of 42\%.

In the patent domain, there are three models trained on patent text, namely PatentBERT by \textcite{lee_patent_2020}, Google’s BERT for patents \parencite{srebrovic_leveraging_2020} and the SEARCHFORMER (not publicly available) by \textcite{vowinckel_searchformer_2023}.

Because BERT models were trained to predict masked tokens and the next sentence, they do not perform effectively at identifying similar documents. This is a necessary prerequisite to identify potential knowledge flows between documents. The SPECTER model \parencite{cohan_specter_2020} and its successor SPECTER2 \parencite{singh_scirepeval_2023} address these limitations through citation-informed learning. The difference between SPECTER and SPECTER2 is simply the underlying training dataset: At the time of training of SPECTER, Semantic Scholar consisted only of biomedical publications and those from computer science. SPECTER2 uses a more recent version of Semantic Scholar, encompassing all scientific disciplines. More specifically, the authors fine-tuned SciBERT via contrastive learning between two publications, an actually cited publication and an uncited random publication.

However, SPECTER was only trained on scientific publications and the language specific to patents likely differs from that relevant to scientific publications. Recently, \textcite{ghosh_paecter_2024} develop the PaECTER model. It was trained in a similar way as SPECTER/SPECTER2 as it leverages credible citation information between patents. Credible citation information refers to citations added by examiners at the European Patent Office (EPO).

Thus we strive to fill the gap by presenting a rigorously trained and tested cross-corpus language model.

\section{Data}

For training and the evaluations, we use the \textit{Reliance on Science} dataset provided by \textcite{marx_reliance_2020,marx_reliance_2022}. Both of these data sets are rigorously curated, lending a high degree of reliability and applicability. The dataset contains two parts: The Patent-Paper Pairs (PPP) and Patent-to-Paper Citations (PPC). All datasets contain the patent publication number and the OpenAlex IDs of the publications. OpenAlex is a comprehensive, community-curated database of scholarly works, authors, institutions, and more \parencite{priem_openalex_2022}.

The PPP data links USPTO patents to publications that are about the same invention. To qualify, a publication must be co-authored by at least one inventor and published within 9 years around the patent publication. Matches are then verified. The PPP dataset provides four confidence scores based on how certain algorithm and human judgment are that the reference and the scientific publication belong together.\footnote{The cutoff values are 0.99 for category 4 (very high), 0.9 for category 3 (high), 0.8 for category 2 (medium), and 0.7 for category 1 (low).}

The PPC data set is a patent data set with all the cited scientific publications for each patent. The data set contains around 47 million citation links for about 7 million unique patents. It includes several authorities, including USPTO (approx. 34 million), EPO (5.7 million), WIPO (4.1 million). The PPC data contains confidence scores ranging from 1 (lowest) to 10 (highest).

In some cases, the PPC and PPP overlap. Yet in most of the cases, there are no direct citations between the patent and the paper or vice versa. In the most obvious case, this would not be possible. If a patent cites a publication on the same invention, it would destroy the novelty. However, there are many publications with an earlier publication date than the patent, due to the grace period. In these cases, the publication already exists but is not cited by the patent.

\section{Fine-tuning SPECTER2 and PaECTER}

Hitherto, in innovation studies, there exist only language models specific to one corpus: either publications or patents. Thus we develop two cross-corpus models, Pat-SPECTER and Pub-PaECTER.

Pat-SPECTER and Pub-PaECTER are derivatives of SPECTER and PaECTER respectively. Pat-SPECTER is the SPECTER2 model fine-tuned on the training data set for the PaECTER, and the Pub-PaECTER is the original PaECTER fined-tuned on the training data set of the SPECTER.

\subsection{Training Data}
The training dataset of SPECTER and SPECTER2 are constructed in the same way: For each focal document, they use five triples, where the second element is a cited document (called "positive") and the third element is a non-cited document (called "negative"). Negatives are furthermore separated into easy negatives and hard negatives. Easy negatives are randomly selected uncited documents, while hard negatives are indirectly cited documents (publications cited by cited publications but not by the focal document).

The training dataset of PaECTER \parencite{ghosh_paecter_2024} is akin to the SPECTER/SPECTER2 approach: For each of the 300k patent documents, there are also five triplets with positives and negatives. Positives are patents cited with citation category X, Y, I and A. Easy negatives are patents sharing at least some CPC classes, while hard negatives are indirectly cited patents. Crucial for the analysis, only EPO patents are considered, because at the EPO references to other patents are solely added by examiners, and with specific citation categories.

\subsection{Training}
Equipped with these models and datasets, we fine-tune them in a cross-corpus manner: First, we fine-tune SPECTER2 (base) on the training dataset of PaECTER. The resulting model is called \textit{Pat-SPECTER}. Second, we fine-tune PaECTER on the training dataset of SPECTER. The resulting model is called \textit{Pub-PaECTER}. Both models are trained on 4 NVIDIA A100-SXM4-40GB GPUs using \textit{Decoupled Weight Decay Regularization} \parencite{loshchilov_decoupled_2019} with a triplet loss margin of 1.

\begin{figure}[H]
    \caption{Overview of the models used in the comparisons\label{fig:deliverables}}
    \centering
    \includegraphics[width=15cm]{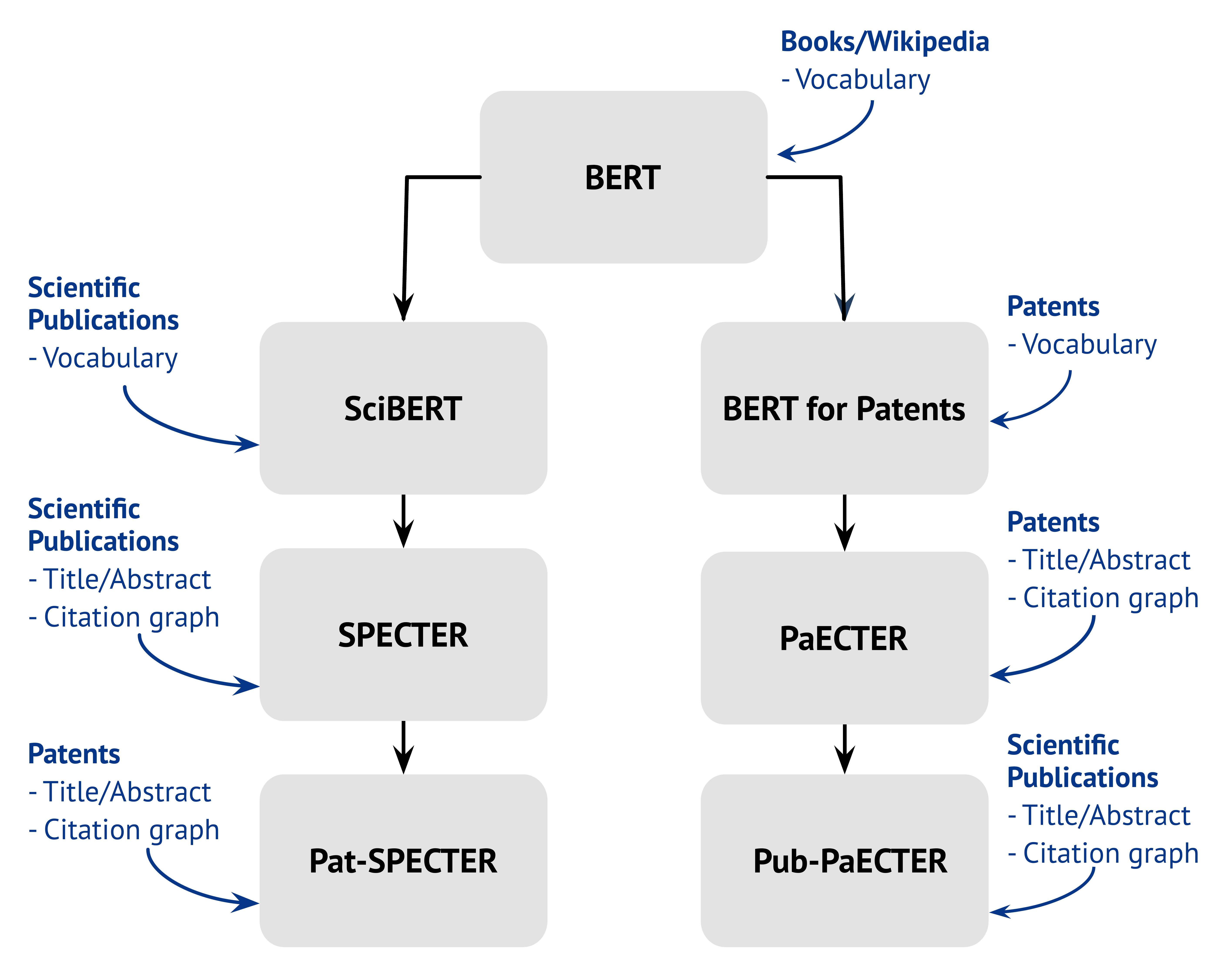}

    \raggedright \textbf{Notes:} BERT \parencite{devlin_bert_2019} is a general language model for the English language trained on a variety of public corpora. SciBERT \parencite{beltagy_scibert_2019} is a BERT with a vocabulary specific to scientific papers (biomedical and computer science). BERT for Patents \parencite{srebrovic_leveraging_2020} is a BERT with a vocabulary specific to patents. SPECTER \parencite{cohan_specter_2020} augments the SciBERT model using paper citations. PaECTER \parencite{ghosh_paecter_2024} augments the BERT for Patents model using patent citations. Pat-SPECTER and Pub-PaECTER are developments reported in this paper.
\end{figure}

\textit{Pat-SPECTER} is trained for one epoch with a per-device batch size of 8 and a learning rate of $1e$--$5$. This requires about 19 hours per epoch with validation every 2000 steps. \textit{Pub-PaECTER} is trained for two epochs with a per-device batch size of 4 and gradient accumulation of 4, a learning rate of $5e$--$6$, and the same validation schedule. This takes about 18 hours per epoch. The number of epochs for both models was determined based on the observation that further training degraded validation performance.
All fine-tuning was performed at the Max Planck Computing and Data Facility (MPCDF). Figure \ref{fig:deliverables} illustrates how the models relate to each other.

\section{Finding a cross-corpus language model\label{ch:horse_race}}

\subsection{Experimental setup}
Our goal is to assess which similarity models trace knowledge flows from science to technology best. We compare the following models against each other: BERT, SciBERT, BERT for Patents, SPECTER, SPECTER2, PaECTER, Pat-SPECTER, Pub-PaECTER. For comparison we also add BM25.

For each of these models, the task is to rank 5 related publications out of 30 as high as possible. One can think of this task as a citation recommendation where a potential examiner seeks to read related non-patent literature. A good recommendation machine is one that suggests only relevant publications.

As ground-truth data, we use EPO patents with at least 5 certain references to different NPL. We derive this data from the Patent-Paper-Citation dataset but conduct the analysis at the DocDB family level. That is, we deduplicate patents at the family level and use the earliest publication year for all patents of the family. If the family cites more than 5 papers, we randomly select 5 of these, which we call "positives". These are contrasted with non-cited papers called "negatives". We only consider patent-paper citations of the highest confidence, that is score 10. Overall, we find 52,696 DocDB families satisfying the criteria.

We derive titles and abstracts of patents from PATSTAT Autumn 2024, and the titles and abstracts of publications from OpenAlex and Scopus.\footnote{For several publishers, OpenAlex does not provide a work's abstract. We attempt to replenish the abstract from Scopus via the work's DOI.} We only consider English patents and publications. In some cases, OpenAlex does not provide a language flag, in which case we estimate the language ourselves; in other cases, the language flag is incorrect, but we choose not to do anything about this.

\begin{figure}
    \centering
    \caption{KDE of the time lag between citing patent and cited paper\label{fig:kde_citation-lag}}
    \includegraphics[width=\linewidth]{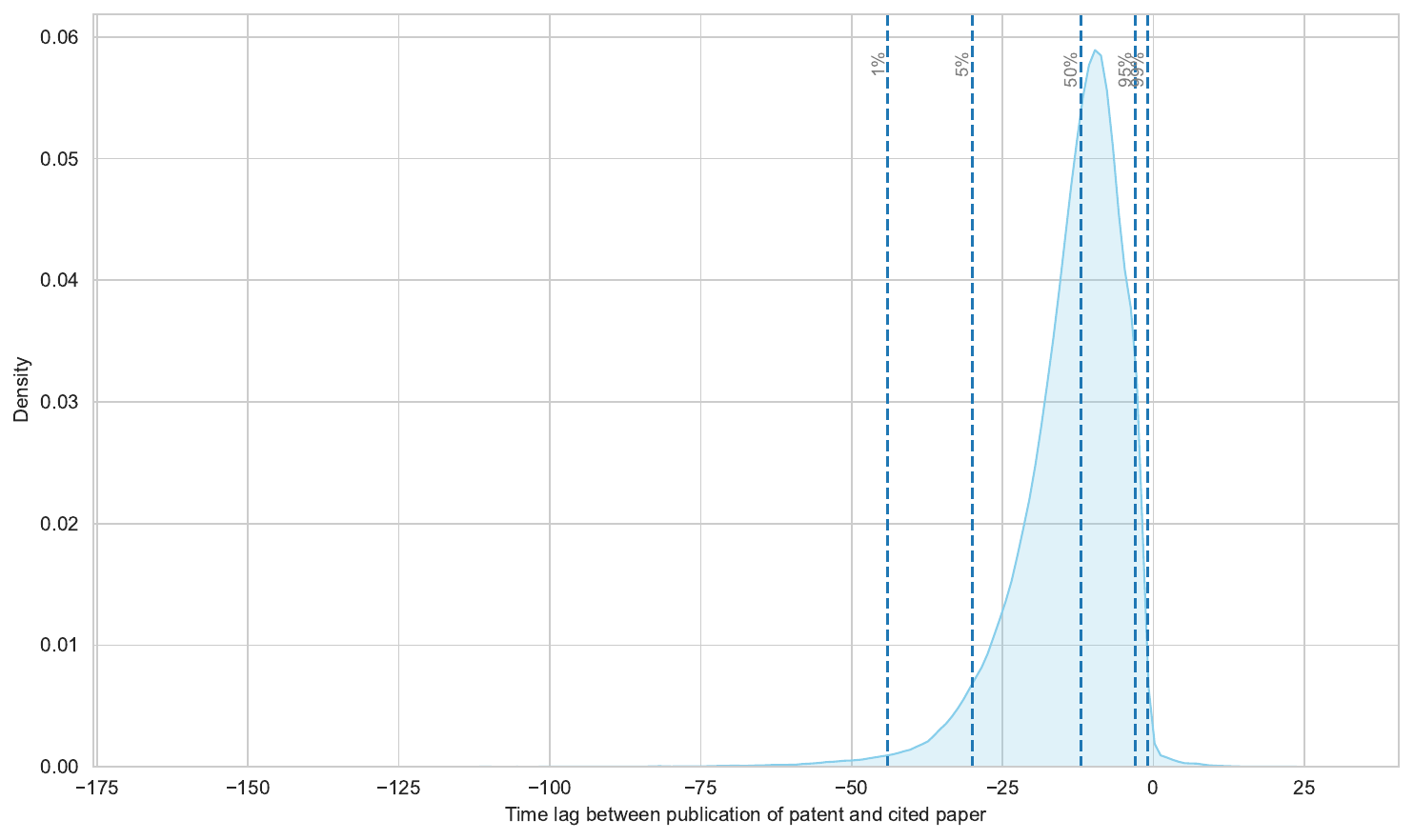}
    \raggedright \textbf{Notes:} This Kernel-Density Estimation plots the distribution of the time lag in years between the publication date of a cited paper and the publication date of the citing patent. Only papers with English abstract and valid publication date considered.
\end{figure}

In total, we find 521,481 citation links to 273,175 distinct papers. The time lag between the publication date of the paper and that of the patent can be substantial. Figure \ref{fig:kde_citation-lag} shows that the median age is 8 years (meaning, the paper got published 8 years before the patent), while the 1st percentile equals 38 years. In a few thousand cases, the paper was published after the patent.

Because of this citation lag pattern, we restrict the risk set of negative (i.e., not cited) papers to 1 to 38 years prior to the publication date of the patent. Out of this patent-specific risk set, for each patent we sample the 25 negative papers provided an that English abstract exists in either OpenAlex or Scopus.

\subsection{Results}

The test protocol is hence as follows:
\begin{enumerate}\itemsep0pt
    \item The model generates embeddings for all 31 documents in a triplet (the focal patent and the 30 publications).
    \item We calculate the similarity between the focal patent's embeddings and the embeddings of the 30 publications.
    \item The documents are then ranked from most similar to least similar based on the cosine distance.
\end{enumerate}
\noindent This process is repeated for each of the 750,000 triplets.

Each model's goal is to place the actually cited publications as highly as possible, ideally they should be the first five.

We use three standard metrics from Information Retrieval and Machine Learning to assess the quality of ranked lists: Average Rank of First Relevant (RFR), Mean Average Precision (MAP), and Mean Reciprocal Rank at 10 (MRR@10). RFR is the simplest, denoting the position in the ranked list at which the first relevant item appears. Lower values indicate better performance, as relevant items appear earlier. The metric is then averaged over all triplets.

Mean Average Precision (MAP) averages the precision scores across all relevant items in the ranking for each query and then averages across all queries $q \in Q = 1000$ according to formula~\eqref{eq:map}:

\begin{equation}\label{eq:map}
    \text{MAP} = \frac{1}{Q} \sum_{q=1}^{Q} \left( \frac{1}{R_q} \sum_{k \in \mathcal{R}_q} \text{Precision}@k \right)
\end{equation}

\noindent Here, $R_q = 5 \forall q$, while Precision$@k$ denotes the proportion of relevant documents among the top $k$, i.e., the proportion of relevant documents in the top $k$ results. This metric rewards systems that not only retrieve relevant results but also order them effectively throughout the ranking.

Mean Reciprocal Rank at 10 (MRR@10) computes the average of reciprocal ranks of the first relevant result over all queries, restricted to the top 10 results. For $q \in Q$, the metric is calculated according to formula~\eqref{eq:mrr}:

\begin{equation}\label{eq:mrr}
    \text{MRR@10} = \frac{1}{Q} \sum_{q=1}^{Q} \frac{1}{\text{rank}_q}
\end{equation}

\noindent If no relevant result appears in the top 10, the reciprocal rank is 0 for that query. MRR@10 emphasizes early correct retrieval and is commonly used when a single relevant result is sufficient.

\begin{table}[ht]
  \caption{Rank-Aware Evaluation of Different Models on the Cross-Corpus Dataset\label{tab:cross_corpus_eval}}
  \centering %\begin{tabular}{l cc cc cc}
%    \toprule
%    & \multicolumn{2}{c}{Avg. RFR} & \multicolumn{2}{c}{MAP} & \multicolumn{2}{c}{MRR@10} \\
%    \cmidrule(lr){2-3} \cmidrule(lr){4-5} \cmidrule(lr){6-7} Model & CLS & Mean & CLS & Mean & CLS & Mean \\
%    \midrule
%    BERT & 2.52 & 1.29 & 46.76 & 79.05 & 69.86 & 91.52 \\
%    SciBERT & 2.48 & 1.37 & 49.23 & 71.14 & 71.75 & 90.55 \\
%    BERT for Patents & 1.20 & 1.10 & 78.51 & 85.99 & 93.14 & 96.97 \\
%    SPECTER & 1.08 & 1.13 & \textbf{91.64} & 86.57 & 97.72 & 96.23 \\
%    SPECTER2 & 1.11 & 1.35 & 88.29 & 76.51 & 95.94 & 91.62\\
%    Pat-SPECTER & \textbf{1.05} & 1.12 & 91.38 & 87.24 & \textbf{98.04} & 96.06 \\
%    PaECTER & 1.13 & 1.07 & 86.45 & 89.72 & 96.12 & 97.55 \\
%    Pub-PaECTER & 1.32 & 1.25 & 76.49 & 79.57 & 92.51 & 94.15 \\
%    \bottomrule
%\end{tabular}

\begin{tabular}{l cc cc cc}
    \toprule
    & \multicolumn{2}{c}{Avg. RFR} & \multicolumn{2}{c}{MAP} & \multicolumn{2}{c}{MRR@10} \\
    \cmidrule(lr){2-3} \cmidrule(lr){4-5} \cmidrule(lr){6-7} Model & CLS & Mean & CLS & Mean & CLS & Mean \\
    \midrule
    BERT & 2.51 & 1.28 & 47.13 & 80.83 & 68.96 & 91.80 \\
    SciBERT & 3.70 & 1.73 & 40.69 & 67.88 & 60.78 & 86.32 \\
    GTE-large & 2.28 & 1.07 & 52.79 & 93.23 & 75.75 & 98.39 \\
    BERT for Patents & 1.20 & 1.16 & 78.90 & 86.97 & 93.48 & 95.98 \\
    PatentSBERTa & 1.10 & 1.09 & 88.67 & 88.43 & 97.08 & 96.99 \\
    SPECTER & 1.09 & 1.23 & 91.64 & 84.50 & 97.37 & 94.63 \\
    SPECTER2 & 1.11 & 1.26 & 89.89 & 80.17 & 96.21 & 93.31 \\
    PaECTER & 1.09 & 1.07 & 89.42 & 91.89 & 97.43 & 98.03 \\
    %E5-large-v2 & 1.06 & 1.05 & 90.47 & 92.41 & 97.99 & 98.29 \\
    %BGE-large-en-v1.5 & \textbf{1.04} & \textbf{1.03} & \textbf{95.29} & \textbf{95.98} & \textbf{98.93} & \textbf{99.05} \\
    \addlinespace
    \textbf{Fine-tuned Models} & & & & & & \\
    \cmidrule(lr){1-1} \addlinespace Pat-SPECTER & \textbf{1.06} & 1.12 & \textbf{93.11} & 88.85 & \textbf{98.06} & 96.51 \\
    Pub-PaECTER & 1.44 & 1.37 & 73.02 & 76.56 & 89.73 & 91.65 \\
    \bottomrule
\end{tabular}

  \raggedright \textbf{Notes:} "Avg. RFR" is the average rank of the first relevant (i.e., actually cited) publication. "MAP" is the mean-average precision, which takes into account the precision and the recall at every position where a relevant item appears (equation \eqref{eq:map}). "MRR@10" is the mean reciprocal rank of the first relevant publication within the 10 closest publications (equation \eqref{eq:mrr}). "CLS" and "Mean" refer to different ways we compute embeddings for a text of multiple sentences: "CLS" concatenates all sentences into one sentence while "Mean" takes the average of the sentences' embeddings.
\end{table}

The results in Table \ref{tab:cross_corpus_eval} reveal that Pat-SPECTER performs well in terms of \textit{avg. RFR} and \textit{MRR@10} compared to other models. However, it comes out only second to SPECTER in terms of \textit{MAP} metric, though the difference is not statistically significant (see Table \ref{tab:cross_corpus_stat_significance} in Appendix). 

Since \textit{avg. RFR} and \textit{MRR@10} evaluate the rank of the most similar publication for a given patent, Pat-SPECTER's strong performance in these metrics suggests that it excels at early detection of the most relevant publications. In general, its slightly lower performance in \textit{MAP}, which considers the ranks of all relevant publications, indicates that it may need to retrieve more publications to identify all relevant ones.

Thus we conclude that Pat-SPECTER is the most suitable model for cross-corpus comparison involving scientific publications and patents. The runner-up model is SPECTER, but Pat-SPECTER statistically dominates SPECTER.

\FloatBarrier

\section{Validation Exercises}
To make the Pat-SPECTER model more palpable, we test it on two applications which we detail in this section.

The tasks at hand are these:
\begin{enumerate}\itemsep0pt
    \item Can the Pat-SPECTER distinguish between Patent-Paper Pairs, Patent-Paper Citations and random patent-paper pairing?
    \item Can Pat-SPECTER predict the paper belonging to a Patent Paper Pair?
\end{enumerate}

All applications make use of the Logic Mill system \parencite{erhardt_logic_2024}. Logic Mill is a scalable and openly accessible software system with the goal to identify semantically similar documents either within and between corpora. Currently, Logic Mill provides Pat-SPECTER embeddings for scientific publications (all OpenAlex works with English-language abstract) and patent documents (all DocDB families represented by their English-language representative, provided it exists). Under the hood it uses an ElasticSearch database, which allows for approximate nearest-neighbor searches across the universe of documents.

However, Logic Mill encodes just one patent document per DocDB family. This representative document does not always coincide with the patent provided in the Reliance on Science-dataset. When the DocDB patent family includes a patent document from the EPO, it takes precedence. More specifically, the pecking order within a DocDB family is EP $>$ WO $>$ US $>$ JP $>$ CN $>$ KR $>$ DE $>$ FR $>$ GB $>$ IT $>$ ES $>$ SE $>$ NL $>$ Other. Lower application IDs break ties. The textual similarity should be extremely high however, if the texts are not actually identical.

\subsection{Separating Patent-Paper Pairs from Patent-Paper Citations}

Our first exercise is to compare the semantic similarity between Patent-Paper Pairs (PPPs) and Patent-Paper citations. Since PPPs are about the same invention, we expect that PPPs should be more similar on average than PPCs.

For our analysis, we use all patent-paper pairs and all patent-paper citations, provided an English abstract exists for both the patent and the paper. We source abstracts from the OpenAlex database and replenish with Scopus, if necessary. Patent abstracts originate from PATSTAT Autumn 2025, however we prioritize abstracts of applications over granted patents if available.

Though both PPPs and PPC come with confidence scores, we chose to include all available pairs and do not restrict to say the highest confidence classes. This approach should yield more conservative estimates as the presence of false PPP or PPC biases them downwards.

Some abstracts contain structure elements (such as headings) or contain copyright statements. These elements are typical for certain journals. Since the abstracts appear to be more similar, it may introduce an unwanted bias. We remove these elements with our abstract cleaning tool.\footnote{To create the embeddings, we use the title and abstract and concatenate these with the \texttt{[SEP]} from \texttt{Autotokenizer}.}

% Upfront it was unclear how big consequences the abstract cleaning would be.
% To investigate the actual effect, we took a sample of 50,000 patent paper pairs. 
% Our statistics show that in around $34\%$ of the cases, abstracts are changed in some way. 
% However, the overall effect on the cosine similarities between the patent and the paper is very small. 
% We found a Cohen's D value of $d=0.04$ and a Pearson correlation of $r=0.994$. (Figure \ref{fig:abstract_cleaning})

% \begin{figure}[t]
%   \begin{center}
%     \caption{Differences in cosine similarities before and after abstract cleaning. The dotted red line indicates equal values.}\label{fig:abstract_cleaning}
%     \includegraphics[width=.6\linewidth]{Figures/abstract_cleaing_effects.pdf}
%   \end{center}
% \end{figure}

Given the embeddings obtained with the Pat-SPECTER, we calculate the cosine similarity for each patent-paper pair. Since some PPP are also PPC, we exclude a PPC if it is also a PPP.

%Additionally, we need to adjust the fraction of PPPs compared to the other pair types. Patent-paper-pairs are not so common in the real world; hence, we adjust the fraction of PPPs. If the fraction is too high, the results will be unrealistic compared to the real world. If the fraction is too low, the modeling results will be inaccurate since we have too few positives. In the latter case, whether a single item is found will make a big difference in the results. In our setup, we have chosen for 10\% PPPs and 90\% random and NPL. An iterative approach is used to make sure the resulting data set sizes are still around the 80/20 split. We ensure that the training and test sets do not have any overlapping papers or patents. We end up with a split of 24,035 pairs for training and 5,661 for testing.

\begin{figure}[t]
    \caption{Distributions of Pat-SPECTER cosine similarities by type of pair}\label{fig:kde-paecter-3cat}
    \centering
    \includegraphics[width=1\linewidth]{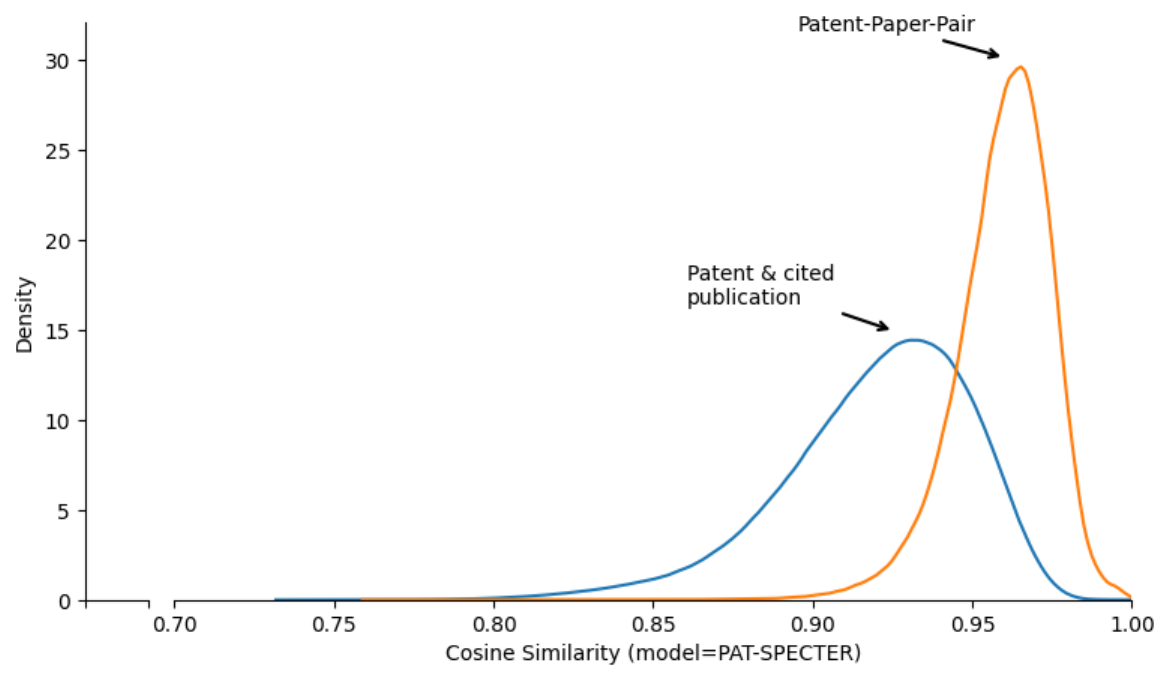}
    \raggedright \textbf{Notes:} These Kernel-Density Estimations plot the distribution of the cosine similarity between Pat-SPECTER embeddings representing a patent and a publication. Only papers with English abstract and valid publication date considered.
\end{figure}

% \begin{figure}[t]
%     \caption{KDE of cosine distributions TEST and TRAIN  non-normalized data calculated with Pat-SPECTER model.}\label{fig:kde-comparison}
%     \centering
%     \includegraphics[width=.9\linewidth]{Figures/93_train_test_KDEs.png}
% \end{figure}

% \begin{figure}[t]
%     \caption{Heatmap of the F1 Scores (with original cosine values)}\label{fig:heatmap}
%     \centering
%     \includegraphics[width=.9\linewidth]{Figures/92_heatmap_results_Unscaled.png}
% \end{figure}

The results in Figure \ref{fig:kde-paecter-3cat} demonstrate a clear separation between the patent-paper pairs and the patent-paper citations. We conclude that Pat-SPECTER is able to reproduce expected semantic similarities. However, the distributions overlap partially.

In practical applications, it is advised to use additional metadata as pre- or post-filtering criteria. For instance, author names could be used to pre-filter documents, refining the search space for identifying PPPs. Within this refined subset, a nearest-neighbor search would be conducted. Documents exceeding the similarity threshold would then be treated as PPPs.

\FloatBarrier

\subsection{Predicting Patent Paper Pairs in the patent-publication universe }
The strength of a nearest-neighbor database is to allow for semantic searches. In this exercise we thus strive to find the paper belonging to a PPP pair. Again, we include all 327,666 patent-paper-pairs where both the patent and the paper are present in Logic Mill.

For each DocDB family with a patent in the PPP dataset, we then retrieve the 1,000 most similar publications according to the Pat-SPECTER model, provided the publication was published within a window of 9 years prior to and 9 years after the patent publication. Then we assess the rank of the actually paired paper. Ideally this should be as low as possible.

\begin{table}
    \caption{Matching statistics for the PPP exercise\label{tab:ppp_rank_distribution}}
    \begin{center}
        \begin{tabular}{lrrrrr}
\toprule
 & \# of obs & Share (in \%) & Median & Mean & Std. dev. \\
Conf. level &  &  &  &  &  \\
\midrule
1 & 62,644 & 63.7 & 58 & 175.8 & 240.9 \\
2 & 74,874 & 69.9 & 44 & 156.4 & 230.2 \\
3 & 113,053 & 79.8 & 26 & 125.4 & 208.0 \\
4 & 77,095 & 89.6 & 7 & 74.0 & 162.5 \\
Total & 327,666 & 75.6 & 26 & 130.0 & 213.7 \\
\bottomrule
\end{tabular}

    \end{center}
    \begin{tablenotes}
        \item \textbf{Notes:} Match rate and rank summary statistics for the paper search for a given US patent among the Patent-Paper pair dataset, by confidence level. Confidence levels relate to algorithmic confidence of the Reliance of Science dataset, with category 4 being the highest confidence.
    \end{tablenotes}
\end{table}

Overall we find the corresponding paper for 251,407 patent-paper pairs among the 1000 most similar candidates. This amounts to a match rate of $\approx$ 75.6\% (Table \ref{tab:ppp_rank_distribution}). Yet, the match rate amounts to as much as 89.6\% for the highest confidence level.

\begin{figure}[ht]
    \caption{Empirical Cumulative Distribution of the rank for the matched PPPs\label{fig:ecdf_ppp-rank}}
    \centering
    \includegraphics[width=0.8\linewidth]{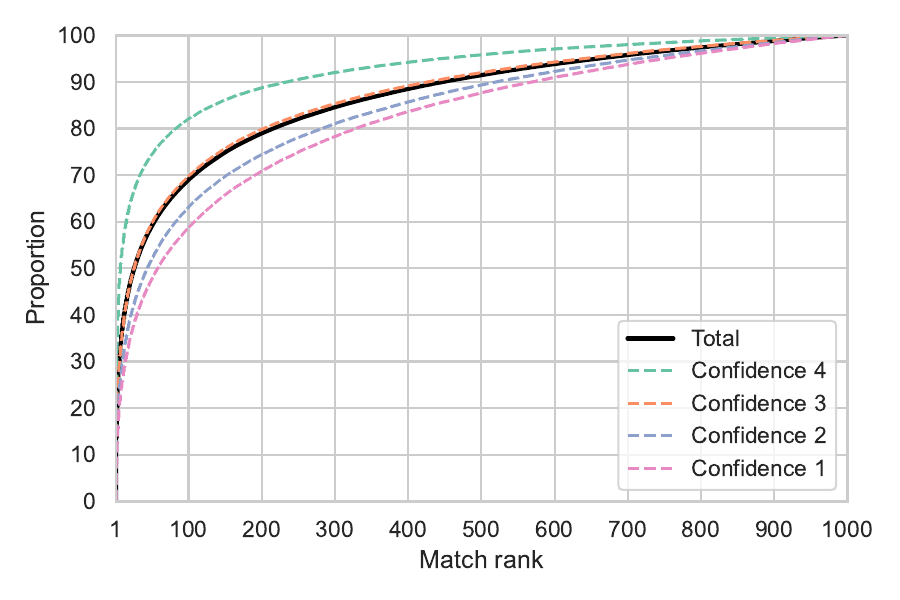}
    
    \raggedright \textbf{Notes:} Empirical cumulative density functions for all matched Patent-Paper pairs, by confidence level. Confidence levels relate to algorithmic confidence of the Reliance of Science dataset, with category 4 being the highest confidence. Table \ref{tab:ppp_rank_distribution} gives mean and median values.
\end{figure}

The ECDF in Figure \ref{fig:ecdf_ppp-rank} displays the match quality for the matched PPPs graphically. The black line represents the total, while the dotted colored lines represent subsets of the data by confidence score. The lines indicate that for most pairs it holds that the matched paper comes early in the distribution. In about 65\% of matched cases, the paper has a rank below 100, and in 90\% of the cases, the true paper ranks below 500. However, the distribution is much denser for the highest confidence.

When we assess the rank of the suggested papers by PPP category, the results become even better. Table \ref{tab:ppp_rank_distribution} gives the corresponding values including the match rate by PPP confidence level.

\FloatBarrier

\subsection{Data limitations\label{sec:data_limitations}}

The most severe limitation stems from lack and incompleteness of our data sources. OpenAlex misses a substantial amount of abstracts. In the May 2024 version, we found that as many as 45.5\% of all works lack an English abstract. Without abstract however, we cannot represent the corresponding document. Thus these documents are exempt from the comparison.

Missing abstracts exist due to many reasons. For instance, abstracts for publications published in journals belonging to Elsevier or Springer Nature are missing, unless they are open access. A second reason is that many works do have an abstract, but not in English. Another reason is the definition of a "work". For OpenAlex, "Works" includes documents published by journals which are not commonly referred to as scientific publications. front matters, back matters, table of contents, advertisements, etc. Even though Logic Mill excludes works labeled as paratext, misclassifications and other errors in OpenAlex perpetuate into Logic Mill.

\FloatBarrier

\section{From Science to Technology}

Having shown that Pat-SPECTER reproduces expected results when it comes to semantic similarity between the universe of English-language patents and English-language publications, we now turn to our research question. Our research question is to study which kind of patents cite unrelated NPL. A paper is considered unrelated if the semantic distance according to the Pat-SPECTER is too large.

We hypothesize that the duty of candor at the USPTO and the Israeli Patent Office results in more unrelated paper citations than at any other authority. The duty of candor requires applicants to disclose (i.e., cite) all related prior art, and violations can render a patent unenforceable. \textcite{cotropia_applicant_2013} document that this duty of candor induces applicants to cite more papers than may be necessary. We add a qualitative perspective to their quantitative finding.

For this task we use all Patent-Paper Citations (PPC) that belong to patents filed between 1970 and 2024. We restrict ourselves to all authorities with at least 10k observations (pairs) in the dataset, excluding WIPO and the Soviet Union. We also drop all pairs where no English abstract for either the cited paper or any member of the patent's DocDB family exists, and we deduplicate patents at the DocDB family level. 

Our final sample thus includes \analysisNFamilies\ patents and \analysisNPapapers\ papers which are connected through \analysisNLinks\ links. 68\% of these have the highest confidence score, and only 3\% have the lowest confidence score.

The dependent variable is whether the actually cited paper is among the 1000 semantically most similar papers published prior to the patent filing. Specifically, given each patent's Pat-SPECTER embedding (or the Pat-SPECTER embedding of the DocDB family's representative document), we perform a $k$-nearest neighbor search for all OpenAlex works published in or before the year of patent filing, with $k=3000$

\begin{figure}
    \caption[Patent-Publication-Ranking]{Distribution of the share of matched paper by patent\label{fig:hist_matching-rate}}
    \centering
    \includegraphics[width=\linewidth]{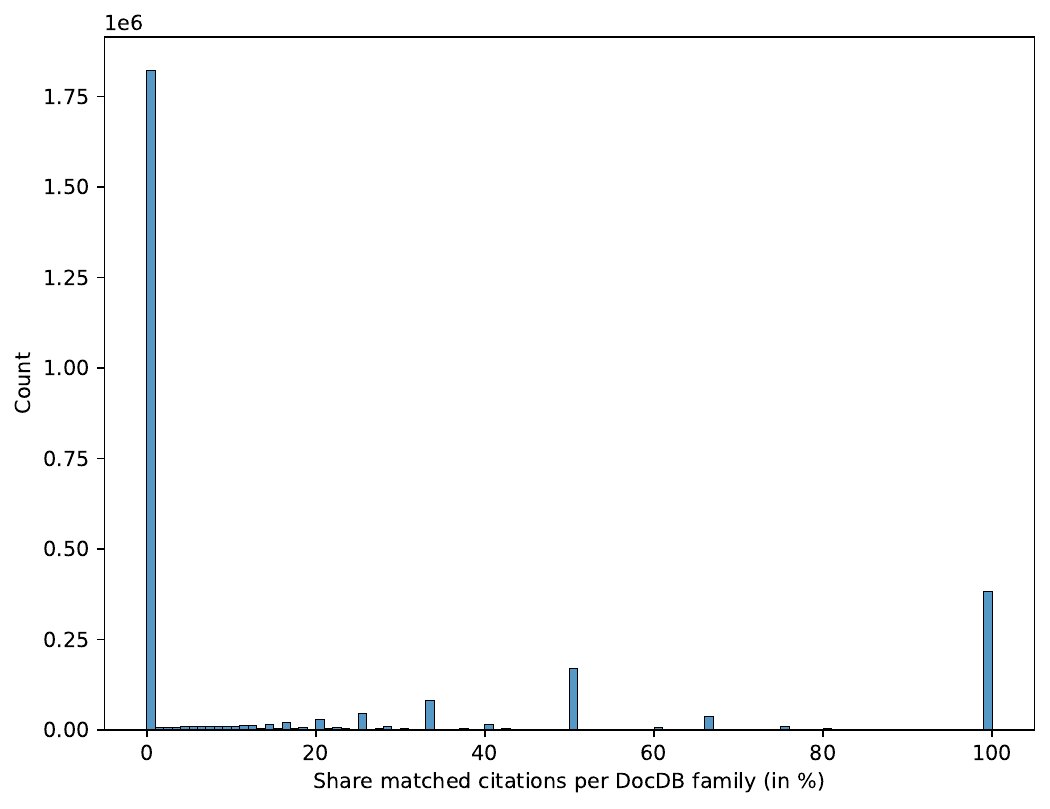}
    
\end{figure}

Out of \analysisNLinks\ potential links, we find the cited paper among the 1k most similar documents in \analysisNMatch\ cases (13\%). Thus, in the majority of cases, cited papers are not specifically similar. Figure \ref{fig:hist_matching-rate} shows share of matched citations per patent, since many patents cite multiple papers.

To explain why some PPC are semantically similar and others not, we run a simple regression of the following form:

\begin{equation}\label{eq:ols_match}
    \mathbbm{1} \{\text{rank}_p(c) < 1000 \} = \alpha + \beta_1 A_p + \beta_2 l_{c, p} + \beta_3 \mathbf{M}_{p} + \gamma s_{c, p} + \tau_p + \epsilon_{p}
\end{equation}

where for every citation $c$ of a patent $p$, we model whether its similarity rank is less than 1000, in relation to the authority $A$, the citation location $l$ and the CPC matrix $M$. The reference authority is the EPO. We control for the confidence score $s_{c, p}$, the patent's number of citations to papers, and the filing year $\tau$ of the patent. Because of the high dimensionality, we have to resort to a linear model (OLS).

Following the above hypothesis, we expect that USPTO and Israeli patents will have a lower matching rate (i.e., the dependent variable will be 0 more often) because of its duty of candor.

\begin{table}[]
    \caption{Regression results for whether the cited paper is among the 1000 most similar documents\label{tab:reg_matched}}
    \begin{center}
        \scriptsize
        \begin{tabular}{llllll}
\toprule
 & \multicolumn{1}{c}{(1)} & \multicolumn{1}{c}{(2)} & \multicolumn{1}{c}{(3)} & \multicolumn{1}{c}{(4)} & \multicolumn{1}{c}{(5)} \\
\midrule
Intercept & 0.16*** & 0.17*** & 0.25*** & 0.20*** & 0.30*** \\
 & (0.00) & (0.00) & (0.00) & (0.00) & (0.00) \\
AU & -0.05*** & -0.06*** & -0.01*** & -0.04*** & 0.00 \\
 & (0.00) & (0.00) & (0.00) & (0.00) & (0.00) \\
CA & -0.11*** & -0.11*** & -0.08*** & -0.10*** & -0.06*** \\
 & (0.00) & (0.00) & (0.00) & (0.00) & (0.00) \\
CN & 0.08*** & 0.07*** & 0.12*** & 0.06*** & 0.09*** \\
 & (0.00) & (0.00) & (0.00) & (0.00) & (0.00) \\
DE & 0.02*** & 0.02*** & 0.03*** & 0.01*** & 0.01*** \\
 & (0.00) & (0.00) & (0.00) & (0.00) & (0.00) \\
EA & 0.14*** & 0.13*** & 0.15*** & 0.15*** & 0.16*** \\
 & (0.01) & (0.01) & (0.01) & (0.01) & (0.01) \\
ES & 0.07*** & 0.07*** & 0.09*** & 0.07*** & 0.09*** \\
 & (0.00) & (0.00) & (0.00) & (0.00) & (0.00) \\
FR & 0.08*** & 0.07*** & 0.07*** & 0.06*** & 0.04*** \\
 & (0.00) & (0.00) & (0.00) & (0.00) & (0.00) \\
GB & 0.10*** & 0.09*** & 0.12*** & 0.09*** & 0.11*** \\
 & (0.00) & (0.00) & (0.00) & (0.00) & (0.00) \\
IL & -0.12*** & -0.12*** & -0.05*** & -0.09*** & -0.03*** \\
 & (0.00) & (0.00) & (0.00) & (0.00) & (0.00) \\
IT & 0.06*** & 0.05*** & 0.08*** & 0.05*** & 0.07*** \\
 & (0.00) & (0.00) & (0.00) & (0.00) & (0.00) \\
JP & -0.06*** & -0.06*** & -0.02*** & -0.06*** & -0.02*** \\
 & (0.00) & (0.00) & (0.00) & (0.00) & (0.00) \\
KR & 0.02*** & 0.01*** & 0.05*** & 0.00*** & 0.04*** \\
 & (0.00) & (0.00) & (0.00) & (0.00) & (0.00) \\
RU & 0.09*** & 0.08*** & 0.11*** & 0.07*** & 0.09*** \\
 & (0.00) & (0.00) & (0.00) & (0.00) & (0.00) \\
TW & -0.06*** & -0.06*** & -0.01*** & -0.06*** & -0.01*** \\
 & (0.00) & (0.00) & (0.00) & (0.00) & (0.00) \\
US & -0.06*** & -0.07*** & -0.05*** & -0.07*** & -0.07*** \\
 & (0.00) & (0.00) & (0.00) & (0.00) & (0.00) \\
Front and body citation & 0.06*** & 0.05*** & 0.05*** & 0.06*** & 0.05*** \\
 & (0.00) & (0.00) & (0.00) & (0.00) & (0.00) \\
Self-citation & 0.14*** & 0.14*** & 0.14*** & 0.14*** & 0.13*** \\
 & (0.00) & (0.00) & (0.00) & (0.00) & (0.00) \\
\midrule
Appln\_Filing\_Year FE &  &  & \multicolumn{1}{c}{\checkmark} &  & \multicolumn{1}{c}{\checkmark} \\
CPC dummies &  &  &  & \multicolumn{1}{c}{\checkmark} & \multicolumn{1}{c}{\checkmark} \\
Confscore FE &  & \multicolumn{1}{c}{\checkmark} &  &  & \multicolumn{1}{c}{\checkmark} \\
N & 15,996,077 & 15,996,077 & 15,996,077 & 15,996,077 & 15,996,077 \\
BIC & 10612621.65 & 10545501.14 & 10463810.85 & 10507507.71 & 10267803.19 \\
Log-Likelihood & -5306161.53 & -5272526.64 & -5231316.56 & -5253529.92 & -5133163.43 \\
\bottomrule
\end{tabular}
    \end{center}
    \begin{tablenotes}
        \item \textbf{Notes:} OLS estimates corresponding to model \eqref{eq:ols_match}. Standard errors in parenthesis. "Self-citation" equals 1 if one of the inventors is author of the cited paper. "Front and body citation" equals 1 if the paper is cited in the front text and the body of a patent.\newline 
* p$<$.1, ** p$<$.05, ***p$<$.01
    \end{tablenotes}
\end{table}

\begin{figure}
    \centering
    
    \caption{Coefficient plot for authorities}
    \label{fig:coefplot_authorities}
    \includegraphics[width=\linewidth]{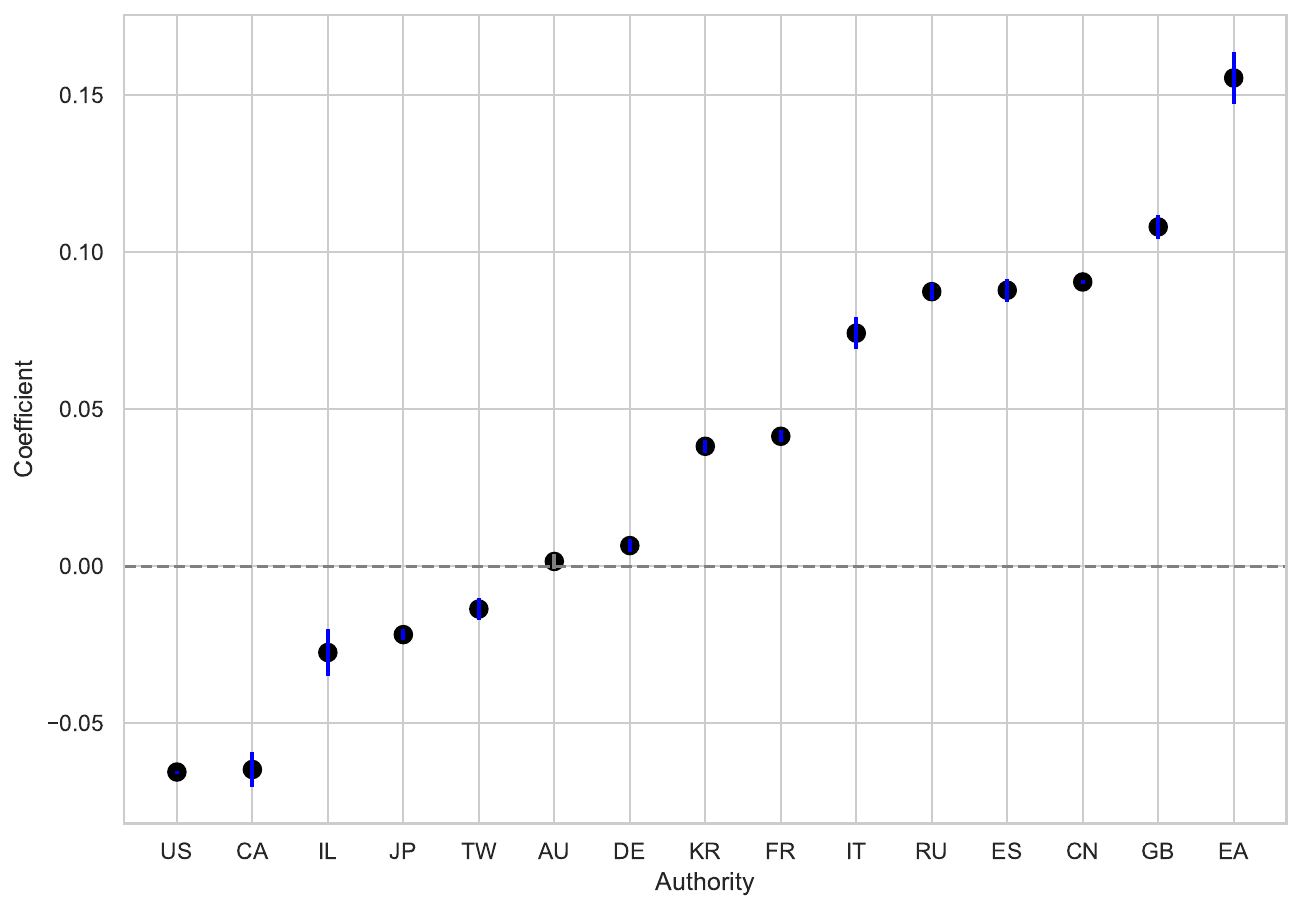}
    \raggedright \textbf{Notes:} This coefficient plot shows the estimated coefficients for the "authority" variable for model \eqref{eq:ols_match} relative to the EPO. Numerical values are in column (5) of Table \eqref{eq:ols_match}. Coefficients statistically significant with $p < 0.1$ highlighted in blue.
\end{figure}

Table \ref{tab:reg_matched} shows the results. We enter control variables one-by-one with column (5) being the most saturated one. Figure \ref{fig:coefplot_authorities} presents the coefficients for the authorities relative to the EPO graphically. All else equal, patents from several authorities are more likely to cite semantically similar publications, namely: Germany, Korea, France, Italy, Russia, Spain, China, Great Britain, and the Eurasian Patent Office. Likewise, compared to the EPO and all else equal, patents filed at the patent authorities of Taiwan, Japan, Israel, Canada and the US are less likely to cite semantically related publications. This finding holds even without all controls (column (1)).

The finding for US and IL patents confirms our hypothesis: Publications cited in US or IL patents are 7\% less likely to be among the 1000 most similar ones as compared to the EPO, even for the same CPC classes. We thus conclude that Pat-SPECTER produces results that show that the system works.

\begin{figure}
    \centering
    
    \caption{Coefficient plot for classes}
    \label{fig:coefplot_classes}
    \includegraphics[width=\linewidth]{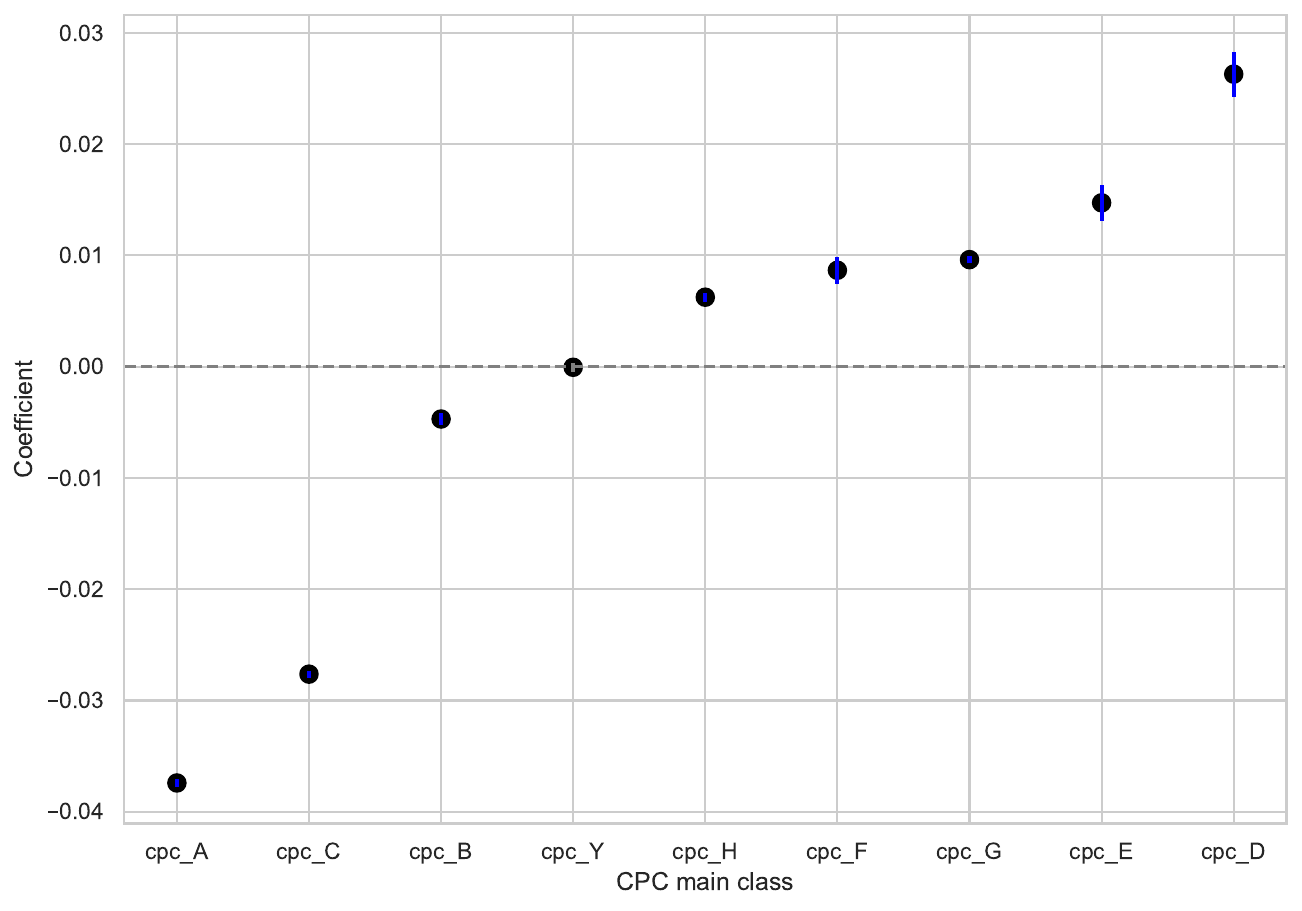}
    \raggedright \textbf{Notes:} This coefficient plot shows the estimated coefficients for the "CPC" variable for model \eqref{eq:ols_match}. Numerical values are omitted from Table \eqref{eq:ols_match}. Coefficients statistically significant with $p < 0.1$ highlighted in blue.
\end{figure}

Finally, in Figure \ref{fig:coefplot_classes} we present the coefficients for the CPC matrix from equation \eqref{eq:ols_match}. There is no reference CPC class since they are not mutually exclusive. We find that patents linked to CPC classes A (Human necessities), C (Chemistry; metallurgy) or B (Performing operations; transporting) are less likely to cite semantically similar publications than patents not linked to these CPC classes. On the other hand, patents linked to CPC classes H (Electricity), F (Mechanical engineering; lighting; heating; weapons; blasting engines or pumps), G (Physics), E, (Fixed constructions) and D (Textiles; paper) are more likely to cite semantically similar publications. Only the heterogeneous residual category Y (for new technological developments) displays a precise 0 coefficient.

% Table \ref{tab:pat_pub_map_mrr} presents the MRR (eq. \eqref{eq:mrr}) and MAP (eq. \eqref{eq:mrr}) for various $k \in \{5,10,20,50,100\}$. The results indicate that for patents with a truly cited paper among the top 5, users find a relevant item at position $1/0.16 \approx 6$ on average. This considers a universe of currently 140 million available documents.

% \begin{table}
%     \caption{MRR and MAP for patent-to-publication citations\label{tab:pat_pub_map_mrr}}
%     \begin{center}
%         \input{Tables/evaluation_patent2publications}
%     \end{center}
%     \begin{tablenotes}
%         \item \textbf{Notes:} Table shows the Mean Reciprocal Rank (MRR) and Mean Average Rank (MAP) at different rank cutoffs $k$.
%     \end{tablenotes}
% \end{table}

% To understand the omission of a significant share of citations, we manually examined cases where matches were not found within our approximate nearest neighbor (ANN) results. Our analysis reveals that many of these citations were applicant-added rather than examiner-cited. These applicant-added citations often appear in the patent description, a context not considered during model training nor encoded in our system.

\FloatBarrier

\subsection{Interpretability}
Pat-SPECTER, like SPECTER2 and PaECTER, is the result of contrastive learning. Contrastive learning models produce embeddings whose cosine similarities are highly similar. In our case, they are clustered around 0.86, while we never observe minimum cosine similarities smaller than 0.7.
% TODO: Find reference for above claim

Hence, we caution against the interpretation of the absolute cosine similarities of two Pat-SPECTER embeddings. The model was trained to produced ordinal rankings, and answers question like: Which document is more similar to my focal document than others? It is not trained state that a document is twice as similar to a focal document than another one. 

\section{Conclusion}
As transformer-based language and similarity models are trained on a single corpus (such as patents), they often perform poorly on other corpora (such as publications). We train  multiple self-trained cross-corpus language models and benchmark them on the same prediction task.

The winning language model, Pat-SPECTER, demonstrates promising performance in finding semantically related publications for a given patent, and vice versa. It uses the latest advances in machine learning, makes efficient use of all textual information (within the 512-token limit), is publicly available and produces replicable results.

The Pat-SPECTER can thus link documents in the technology domain to documents in the science domain in the absence of citations, or in the presence of strategic citations. As such, it can be helpful in many analyses at the heart of innovation economics. These include knowledge transfer, direction of technological and scientific development, and prior art search.

As many of these analyses require comparisons against the entire corpus, we recommend using the Logic Mill system \parencite{erhardt_logic_2024}. It stores Pat-SPECTER embeddings for all DocDB patent families (with an English-language member) and all OpenAlex works (with an English-language abstract) for fast and convenient (approximate) Nearest Neighbor searches. We hope this may overcome the search problems that inventors face when engaging with academic literature \parencite{bikard_made_2018}.

Future analyses might experiment with representation of other patent parts, such as claims, and the incorporation of patent metadata, such as CPC classes.

% BIBLIOGRAPHY
\printbibliography[heading=bibintoc]

% APPENDIX
\newpage
\section{Appendix: Statistical Significance Test Pat-SPECTER}
\begin{table}[ht]
    \centering
    \caption{Statistical Significance Testing of Rank-Aware Evaluation of Pat-SPECTER versus Different Models on a Cross-Corpus Dataset\label{tab:cross_corpus_stat_significance}}
    \begin{tabular}{lccc}
\toprule
 & Avg. RFR & MAP & MRR@10 \\
\midrule
BERT & 0.214\sym{***} (0.007) & -0.123\sym{***} (0.001) & -0.063\sym{***} (0.001) \\
\addlinespace
SciBERT & 0.668\sym{***} (0.007) & -0.252\sym{***} (0.001) & -0.117\sym{***} (0.001) \\
\addlinespace
BERT for Patents & 0.095\sym{***} (0.007) & -0.061\sym{***} (0.001) & -0.021\sym{***} (0.001) \\
\addlinespace
SPECTER & 0.028\sym{***} (0.007) & -0.015\sym{***} (0.001) & -0.007\sym{***} (0.001) \\
\addlinespace
SPECTER2 & 0.049\sym{***} (0.007) & -0.032\sym{***} (0.001) & -0.019\sym{***} (0.001) \\
\addlinespace
PaECTER & 0.006 (0.007) & -0.012\sym{***} (0.001) & -0.000 (0.001) \\
\addlinespace
Pub-PaECTER & 0.304\sym{***} (0.007) & -0.166\sym{***} (0.001) & -0.064\sym{***} (0.001) \\
\addlinespace
Constant & 1.063\sym{***} (0.005) & 0.931\sym{***} (0.001) & 0.981\sym{***} (0.001) \\
\midrule
Adjusted $R^2$ & 0.034 & 0.180 & 0.044 \\
Observations & 421,568 & 421,568 & 421,568 \\
\bottomrule
\end{tabular}

    \RaggedRight \footnotesize \textbf{Notes:} Standard errors in parentheses. Base level model is Pat-SPECTER. All models use their best pooling method, as derived from Table \ref{tab:cross_corpus_eval}.\\
    \sym{*} \(p<0.05\), \sym{**} \(p<0.01\), \sym{***} \(p<0.001\)
\end{table}

% \subsection{Paper-to-Paper citations}

% We perform another test case where the goal is to predict citations among publications.

% Our evaluation set is the test set of SPECTER with 996 triplets (see section \ref{sec:specter_pubdata}). 
% Every record contains a focal publication, five cited publications (positives), and 25 random publications (negatives).
% % This test is specifically designed to evaluate the Pub-PaECTER model for predicting citations among publications using the SPECTER test dataset.

% \begin{table}[ht]
%     \centering
%     \caption{Rank-Aware Evaluation of Different Models on SPECTER Test Dataset\label{tab:pubpaecter_eval}}
%     \vspace*{5mm}
%     \input{Tables/pubpaecter_eval}
% \end{table}

% Table \ref{tab:pubpaecter_eval} shows that Pub-PaECTER performs poorly in predicting citations within scientific articles, while SPECTER2 consistently exhibits the best performance. 
% This suggests that Pat-SPECTER fails to capture the notion of similarity and dissimilarity between publications, though its performance is only slightly worse than SPECTER2.

\end{document}